\documentclass{article}
\usepackage{PRIMEarxiv}
\usepackage[utf8]{inputenc} % allow utf-8 input
\usepackage[T1]{fontenc}    % use 8-bit T1 fonts
\usepackage{hyperref}       % hyperlinks
\usepackage{url}            % simple URL typesetting
\usepackage{booktabs}       % professional-quality tables
\usepackage{amsfonts}       % blackboard math symbols
\usepackage{amsmath}
\usepackage{amssymb}
\usepackage{nicefrac}       % compact symbols for 1/2, etc.
\usepackage{microtype}      % microtypography
\usepackage{lipsum}
\usepackage{fancyhdr}       % header
\usepackage{graphicx}       % graphics
\usepackage{enumitem}
\usepackage{float}
\usepackage{natbib}
\usepackage{caption}
\usepackage{footnote}
\graphicspath{{media/}}     % organize your images and other figures under media/ folder

\title{Common Sense Is All You Need}

\author{
  Hugo Latapie \\
  \texttt{hmlatapie@gmail.com} \\
}

\begin{document}
\maketitle

\nocite{*}

\begin{abstract}
Artificial intelligence (AI) has made significant strides in recent years, yet it continues to struggle with a fundamental aspect of cognition present in all animals: \textbf{common sense}. Current AI systems, including those designed for complex tasks like autonomous driving, problem-solving challenges such as the Abstraction and Reasoning Corpus (ARC), and conversational benchmarks like the Turing Test, often lack the ability to adapt to new situations without extensive prior knowledge. This manuscript argues that integrating common sense into AI systems is essential for achieving true autonomy and unlocking the full societal and commercial value of AI.

We propose a shift in the order of knowledge acquisition (\textit{ordo cognoscendi}), emphasizing the importance of developing AI systems that start from minimal prior knowledge and are capable of contextual learning, adaptive reasoning, and embodiment---even within abstract domains. Additionally, we highlight the need to \textbf{rethink the AI software stack} to address this foundational challenge. Without common sense, AI systems may never reach true autonomy, instead exhibiting asymptotic performance that approaches theoretical ideals like AIXI but remains unattainable in practice due to infinite resource and computation requirements.

While scaling AI models and passing benchmarks like the Turing Test have brought significant advancements in applications that do not require autonomy, these approaches alone are insufficient to achieve autonomous AI with common sense. By redefining existing benchmarks and challenges to enforce constraints that require genuine common sense, and by broadening our understanding of embodiment to include both physical and abstract domains, we can encourage the development of AI systems better equipped to handle the complexities of real-world and abstract environments. This approach aligns with the ultimate goals of AI research and ensures that investments contribute to sustainable and meaningful progress.
\end{abstract}

% keywords can be removed

\section{Introduction}

\subsection{Background and Motivation}

Artificial intelligence has achieved remarkable feats, from mastering complex games to enabling voice-activated assistants. However, despite these advancements, AI systems often lack \textbf{common sense}---the ability to understand and reason about the world in a way that all animals do. This deficiency becomes especially evident in autonomous agents operating in dynamic, real-world environments, such as self-driving cars, robotic assistants, and conversational systems, as well as in abstract problem-solving tasks like the Abstraction and Reasoning Corpus (ARC) challenge \footnote{\citet{Chollet2019}}.

For instance, while AI-powered vehicles can navigate predefined routes using extensive sensor data and mapping, they may struggle with unexpected obstacles or novel scenarios that require adaptive decision-making. Similarly, AI systems tackling problem-solving challenges rely heavily on extensive training data, limiting their ability to generalize and reason beyond their programmed knowledge. Even AI systems that pass the Turing Test, capable of engaging in human-like conversation, may lack genuine understanding and common sense reasoning.

The absence of common sense in AI not only hampers performance but also poses safety risks and hinders the achievement of true autonomy. As AI systems become more integrated into society, addressing this gap becomes increasingly critical. \textbf{All animals exhibit common sense necessary for survival}, demonstrating basic intelligence through their interactions with the environment. This observation serves as proof that integrating common sense into intelligent systems is both essential and achievable.

\textbf{Recent Expert Perspectives:}

Yann LeCun, a pioneer in AI and deep learning and a recipient of the \textbf{2018 ACM A.M. Turing Award}---the most prestigious prize in computer science---recently highlighted the limitations of current AI systems by stating that \textbf{``AI systems still lack the general common sense of a cat''}\footnote{\citet{Mims2024}}. While the characterization of AI being ``dumber'' than a cat depends on how intelligence is defined, we concur with the underlying sentiment that cats, and animals in general, possess common sense that current AI lacks. This comparison emphasizes the gap between AI capabilities and the intuitive understanding exhibited even by animals with relatively simple cognitive abilities.

By acknowledging this perspective from one of the foremost experts in the field, we underscore the importance of focusing on common sense in AI development. Despite significant advancements made possible by deep learning---a field substantially shaped by Schmidhuber, LeCun, Geoffrey Hinton, Yoshua Bengio, and many other luminaries---there remains a fundamental gap in AI's ability to replicate the common sense exhibited by animals.

\subsection{The Path to Autonomy Requires Common Sense}

The central thesis of this manuscript is that \textbf{common sense is all you need} for AI to reach true autonomy and functionality comparable to human and animal intelligence. We assert that:

\begin{itemize}[leftmargin=*]
    \item \textbf{Current AI Approaches Are Inadequate:}
    \begin{itemize}
        \item Many AI development workflows lack a focus on integrating common sense, leading to limitations in adaptability and understanding.
        \item AI systems built on these approaches may exhibit performance improvements but ultimately plateau, unable to achieve true autonomy.
    \end{itemize}
    \item \textbf{Asymptotic Behavior Towards Theoretical Ideals:}
    \begin{itemize}
        \item Without common sense, AI systems may approach theoretical constructs like AIXI \footnote{\citet{Legg2007}}---a hypothetical optimal agent---but never truly reach practical autonomy.
        \item AIXI requires infinite computational resources and time, making it unattainable in the real world.
        \item Current AI paths may lead to diminishing returns, demanding ever-increasing resources for marginal gains.
    \end{itemize}
    \item \textbf{Focusing on Common Sense Is Essential:}
    \begin{itemize}
        \item By integrating common sense, AI systems can adapt to new situations, make intuitive decisions, and operate autonomously without exhaustive computational demands.
        \item This focus aligns AI development with the practical capabilities exhibited by humans and animals.
    \end{itemize}
\end{itemize}

\subsection{The Need to Rethink AI Software Architectures}

In recognizing the limitations of current AI systems in achieving true autonomy, it becomes apparent that we may need to \textbf{rethink the entire software stack} used in AI development. Traditional software architectures are often not designed to accommodate the integration of common sense reasoning. This realization is challenging, as it requires a departure from established methodologies and an openness to fundamentally new approaches. However, to develop AI systems capable of reliable and autonomous operation, redesigning the software stack to support common sense integration may be essential.

Our approach questions whether incremental improvements within existing frameworks are sufficient. We propose that achieving true autonomy may necessitate a fundamental redesign of AI software architectures to incorporate mechanisms that enable contextual learning, adaptive reasoning, and embodiment, both physical and abstract. This shift may involve integrating concepts from cognitive science, neuroscience, and other disciplines to build systems that learn and reason in ways more akin to biological intelligence.

\subsection{Objectives of the Manuscript}

The main objectives of this manuscript are to:

\begin{itemize}[leftmargin=*]
    \item \textbf{Define Common Sense in Detail:}
    \begin{itemize}
        \item Provide a comprehensive definition of common sense in the context of AI, emphasizing its components such as contextual learning, adaptability, starting from minimal prior knowledge, and highlighting its presence in all animals.
        \item Introduce a broader concept of embodiment that applies not only to physical interactions but also to abstract domains, such as problem-solving tasks like the ARC challenge.
    \end{itemize}
    \item \textbf{Analyze Current Problems and Approaches:}
    \begin{itemize}
        \item Examine existing benchmarks like the ARC challenge, the Turing Test, and autonomous driving levels to highlight how they fall short in testing and developing common sense in AI systems.
        \item Discuss how current AI development paths, without integrating common sense, may never lead to true autonomy and instead require impractical resources.
    \end{itemize}
    \item \textbf{Incorporate Expert Insights:}
    \begin{itemize}
        \item Reference perspectives from AI leaders like Yann LeCun and Eric Schmidt to underscore the recognized limitations of current AI systems compared to animal intelligence and to address concerns about self-improving AI.
    \end{itemize}
    \item \textbf{Propose a Shift in Development Focus:}
    \begin{itemize}
        \item Argue for the importance of developing common sense-focused variants of existing problems, suggesting that achieving partial success on these more challenging tasks is more valuable than full success on tasks that do not truly test common sense.
        \item Advocate for rethinking the AI software stack to better support the integration of common sense.
    \end{itemize}
    \item \textbf{Address Theoretical Counterarguments:}
    \begin{itemize}
        \item Discuss theoretical challenges, such as the No Free Lunch Theorem, and demonstrate how constraining the problem space to well-defined domains mitigates these limitations.
    \end{itemize}
    \item \textbf{Provide Actionable Recommendations:}
    \begin{itemize}
        \item Offer practical steps for the AI community to redesign benchmarks, develop new evaluation metrics, and embrace interdisciplinary collaboration to prioritize common sense in AI development.
    \end{itemize}
\end{itemize}

\section{Defining Common Sense in AI}

\subsection{Detailed Definition of Common Sense}

\textbf{Common sense} in AI refers to the ability of a system to understand, learn, and reason about the world in a way that is flexible, contextual, and adaptable, much like humans and animals do. Key components include:

\begin{itemize}[leftmargin=*]
    \item \textbf{Contextual Learning:}
    \begin{itemize}
        \item \textit{Definition:} The capacity to interpret and respond to new information based on the context in which it occurs.
        \item \textit{Importance:} Enables AI systems to adjust their behavior in real-time, considering the nuances of each situation.
        \item \textit{Example:} Recognizing that a stop sign is still a stop sign even if partially obscured by foliage or graffiti.
    \end{itemize}
    \item \textbf{Adaptive Reasoning:}
    \begin{itemize}
        \item \textit{Definition:} The ability to modify reasoning strategies when faced with novel or unexpected situations.
        \item \textit{Importance:} Allows AI to handle scenarios not explicitly programmed or encountered during training.
        \item \textit{Example:} Adjusting navigation when a usual route is blocked due to unforeseen circumstances.
    \end{itemize}
    \item \textbf{Embodied Cognition:}
    \begin{itemize}
        \item \textit{Definition:} Understanding that arises from an AI's interactions with environments, which can be physical or abstract, integrating sensory and motor experiences.
        \item \textit{Importance:} Grounds the AI's reasoning in experiences, leading to more intuitive and practical decision-making.
        \item \textit{Example:} In a physical context, learning that slippery surfaces require slower movement to avoid slipping; in an abstract context, recognizing patterns and rules in a problem-solving task like the ARC challenge.
    \end{itemize}
    \item \textbf{Starting from a \textit{Tabula Rasa}:}
    \begin{itemize}
        \item \textit{Definition:} Beginning problem-solving with minimal prior knowledge, relying on fundamental principles and the ability to learn from new information.
        \item \textit{Importance:} Ensures that the AI's reasoning is not confined by overfitting to specific datasets or pre-programmed knowledge.
        \item \textit{Example:} Approaching a completely new game by learning the rules through observation and experimentation rather than relying on prior strategies.
    \end{itemize}
\end{itemize}

By encompassing these components, common sense in AI enables systems to function effectively in complex, dynamic environments---whether physical or abstract---exhibiting flexibility and understanding akin to human and animal cognition.

\subsection{Embodiment Beyond the Physical World}

While embodied cognition traditionally refers to physical interaction with the real world, we propose a broader concept of \textbf{embodiment} that includes interactions within \textbf{abstract or virtual domains}. In this context, embodiment signifies an AI system's ability to engage with any environment---physical or abstract---through perception and action.

\begin{itemize}[leftmargin=*]
    \item \textbf{Cognitive Embodiment:}
    \begin{itemize}
        \item \textit{Definition:} The process by which an AI system interacts with and learns from an abstract environment, such as mathematical problems, puzzles, or logical tasks.
        \item \textit{Example:} In the ARC challenge, an AI system manipulates and interprets abstract patterns and rules to solve tasks, embodying cognition within that constrained domain.
    \end{itemize}
    \item \textbf{Importance of Generalized Embodiment:}
    \begin{itemize}
        \item Recognizes that common sense arises from interaction with structured environments, not solely from physical experiences.
        \item Allows AI systems to develop common sense by learning from a variety of domains, enhancing their adaptability and reasoning capabilities.
    \end{itemize}
\end{itemize}

By embracing both physical and abstract forms of embodiment, we expand the potential for AI systems to acquire common sense across diverse contexts. This generalization is crucial for tasks like the ARC challenge, where physical embodiment is not applicable, but cognitive embodiment plays a significant role.

\subsection{All Animals Exhibit Common Sense}

It is important to recognize that common sense is not exclusive to humans; \textbf{all animals} exhibit common sense essential for their survival. Common sense in animals refers to the innate and learned behaviors that enable them to:

\begin{itemize}[leftmargin=*]
    \item \textbf{Interact Effectively with Their Environment:}
    \begin{itemize}
        \item Animals use their senses to navigate their surroundings, locate food and water, and find shelter.
    \end{itemize}
    \item \textbf{Avoid Danger and Threats:}
    \begin{itemize}
        \item Instinctive responses, such as fleeing from predators or avoiding hazardous environments, demonstrate an understanding of threats.
    \end{itemize}
    \item \textbf{Adapt to Changing Conditions:}
    \begin{itemize}
        \item Animals adjust their behaviors in response to environmental changes, such as seasonal shifts, resource availability, or social dynamics.
    \end{itemize}
    \item \textbf{Reproduce and Ensure Species Continuity:}
    \begin{itemize}
        \item Behaviors related to mating, nurturing offspring, and social cooperation contribute to survival and propagation.
    \end{itemize}
\end{itemize}

\textbf{Expert Perspective Supporting This View:}

Yann LeCun's assertion that \textbf{``AI systems still lack the general common sense of a cat''} underscores the significant gap between AI and even basic animal intelligence\footnote{\citet{Mims2024}}. LeCun, along with Geoffrey Hinton and Yoshua Bengio, received the \textbf{2018 ACM A.M. Turing Award} for their pioneering work in deep learning, which has driven many recent advances in AI. Despite these advancements, LeCun emphasizes that AI systems still lack the common sense that allows animals like cats to navigate and interact effectively with the real world.

\textbf{Implications for AI Development:}

\begin{itemize}[leftmargin=*]
    \item \textbf{Foundation of Intelligence:}
    \begin{itemize}
        \item Since even the simplest animals exhibit common sense necessary for survival, it underscores that common sense is a foundational component of any intelligent agent interacting with the real or abstract world.
    \end{itemize}
    \item \textbf{Achievability in Artificial Systems:}
    \begin{itemize}
        \item If common sense emerges naturally in biological systems through basic interaction with environments, integrating common sense into AI systems is both a necessary and attainable goal.
    \end{itemize}
    \item \textbf{Emphasis on Fundamental Abilities:}
    \begin{itemize}
        \item AI development should prioritize cultivating these fundamental adaptive and responsive behaviors, mirroring the common sense inherent in all animals.
    \end{itemize}
\end{itemize}

% Continue from previous code...

\section{Limitations of Current AI Benchmarks}

\subsection{Analysis of the ARC Challenge}

The \textbf{Abstraction and Reasoning Corpus (ARC)} is a set of tasks designed to evaluate an AI system's ability to generalize and reason abstractly. While it aims to move beyond pattern recognition toward cognitive reasoning, several limitations hinder its effectiveness in testing for true common sense:

\begin{itemize}[leftmargin=*]
    \item \textbf{Use of Prior Knowledge Beyond Assumed Knowledge:}
    \begin{itemize}
        \item \textbf{Current Practice:}
        \begin{itemize}
            \item AI systems often utilize the 400 training problems and may even indirectly access the 400 test problems during development.
        \end{itemize}
        \item \textbf{Issue:}
        \begin{itemize}
            \item There is nothing inherently preventing AI systems from including these problems in their training data, leading to over-reliance on specific examples rather than developing general reasoning abilities.
        \end{itemize}
        \item \textbf{Consequence:}
        \begin{itemize}
            \item AI systems may appear successful on ARC tasks but lack genuine common sense, as they rely on memorized solutions rather than contextual understanding.
        \end{itemize}
    \end{itemize}
    \item \textbf{Not Starting from a Tabula Rasa:}
    \begin{itemize}
        \item \textbf{Problem:}
        \begin{itemize}
            \item The inclusion of extensive prior examples prevents AI systems from approaching problems with minimal preconceptions.
        \end{itemize}
        \item \textbf{Impact:}
        \begin{itemize}
            \item This hinders the evaluation of an AI's ability to reason and adapt to entirely new situations, a key aspect of common sense.
        \end{itemize}
    \end{itemize}
    \item \textbf{Lack of Cognitive Embodiment:}
    \begin{itemize}
        \item \textbf{Observation:}
        \begin{itemize}
            \item While ARC operates in an abstract domain, it requires AI systems to engage in \textbf{cognitive embodiment} by interacting with and learning from the problem environment.
        \end{itemize}
        \item \textbf{Issue:}
        \begin{itemize}
            \item AI systems often fail to embody cognition within this abstract domain, instead relying on pattern matching without deeper understanding or adaptability.
        \end{itemize}
    \end{itemize}
    \item \textbf{Resource-Intensive Solutions with Limited Common Sense:}
    \begin{itemize}
        \item \textbf{Observation:}
        \begin{itemize}
            \item Some AI approaches require extensive computational resources, involving training on vast amounts of data, including unrelated datasets, to achieve impressive but suboptimal results.
        \end{itemize}
        \item \textbf{Human Comparison:}
        \begin{itemize}
            \item Humans can solve ARC problems with ease and higher accuracy, despite not having been explicitly trained on the specific tasks.
        \end{itemize}
        \item \textbf{Implication:}
        \begin{itemize}
            \item The disparity suggests that these AI systems lack the common sense reasoning that allows humans to generalize and adapt quickly.
        \end{itemize}
    \end{itemize}
\end{itemize}

\textbf{Recommendation:}

To truly assess common sense, the ARC challenge should enforce that the \textbf{only} prior knowledge an AI system has is the minimal set specified in the ``assumed knowledge'' section. Additionally, AI systems should be designed to exhibit cognitive embodiment within the abstract domain of ARC, engaging with tasks through contextual learning and adaptive reasoning.

This approach would:

\begin{itemize}[leftmargin=*]
    \item \textbf{Promote Contextual Learning:}
    \begin{itemize}
        \item Encourage AI systems to interpret and solve problems based solely on present information.
    \end{itemize}
    \item \textbf{Reduce Overfitting:}
    \begin{itemize}
        \item Prevent reliance on specific patterns from training data.
    \end{itemize}
    \item \textbf{Enhance Generalization:}
    \begin{itemize}
        \item Better evaluate the AI's ability to apply reasoning to novel situations.
    \end{itemize}
    \item \textbf{Foster Cognitive Embodiment:}
    \begin{itemize}
        \item Enable AI to engage more deeply with abstract problem environments, mirroring how humans approach such tasks.
    \end{itemize}
\end{itemize}

\subsection{Case Study: Full Self-Driving (FSD) and SAE Levels}

The development of \textbf{Full Self-Driving (FSD)} vehicles provides a practical example of the limitations arising from not prioritizing common sense.

\textbf{Understanding SAE Levels of Autonomy:}

\begin{itemize}[leftmargin=*]
    \item \textbf{Level 1 to Level 4:}
    \begin{itemize}
        \item \textbf{Levels 1-3:}
        \begin{itemize}
            \item Involve increasing degrees of automation but still require human intervention in specific circumstances.
        \end{itemize}
        \item \textbf{Level 4:}
        \begin{itemize}
            \item Vehicles can operate without a driver in specific conditions or geofenced areas.
            \item \textbf{May require remote human intervention for edge cases or unexpected situations}.
        \end{itemize}
        \item \textbf{Limitation:}
        \begin{itemize}
            \item The AI systems lack the common sense to handle all possible scenarios independently, thus relying on remote assistance.
        \end{itemize}
    \end{itemize}
    \item \textbf{Level 5 (Full Autonomy):}
    \begin{itemize}
        \item Intended to operate without any human input, remote or otherwise, under all conditions and environments.
        \item \textbf{Challenge:}
        \begin{itemize}
            \item Without integrating common sense, reaching Level 5 is unattainable. AI systems may continue to improve incrementally but will asymptotically approach a ceiling where true autonomy is unachievable due to the lack of common sense reasoning.
        \end{itemize}
    \end{itemize}
\end{itemize}

\textbf{The Role of Common Sense in FSD:}

\begin{itemize}[leftmargin=*]
    \item \textbf{Handling Unpredictable Scenarios:}
    \begin{itemize}
        \item Common sense enables understanding of nuanced contextual cues, such as recognizing atypical human gestures, unexpected obstacles, and complex road conditions.
    \end{itemize}
    \item \textbf{Reducing Dependence on Remote Intervention:}
    \begin{itemize}
        \item AI with common sense can make decisions in edge cases without human assistance, moving closer to true autonomy.
    \end{itemize}
    \item \textbf{Adapting to Novel Situations:}
    \begin{itemize}
        \item With common sense, AI systems can generalize from prior experiences to handle new and unforeseen circumstances.
    \end{itemize}
\end{itemize}

\textbf{``Magic Happens Here'' Phenomenon in FSD:}

\begin{itemize}[leftmargin=*]
    \item \textbf{Workflow Gap:}
    \begin{itemize}
        \item Development processes may assume that incremental improvements will eventually lead to full autonomy without a clear strategy for integrating common sense.
    \end{itemize}
    \item \textbf{Risk:}
    \begin{itemize}
        \item This approach leads to a plateau where AI systems cannot progress beyond Level 4 capabilities, regardless of additional resources or computational power.
    \end{itemize}
\end{itemize}

\textbf{Asymptotic Behavior and Resource Limitations:}

\begin{itemize}[leftmargin=*]
    \item \textbf{Analogy to AIXI:}
    \begin{itemize}
        \item AIXI is a theoretical model of an optimal decision-making agent but is computationally infeasible due to its requirement for infinite resources.
    \end{itemize}
    \item \textbf{Implication:}
    \begin{itemize}
        \item Pursuing autonomy without integrating common sense may result in AI systems that demand exponentially increasing resources for diminishing performance gains.
    \end{itemize}
\end{itemize}

\textbf{Conclusion:}

The lack of common sense in AI systems for FSD illustrates the necessity of addressing this core issue directly. Without doing so, advancements are limited, and true Level 5 autonomy remains out of reach, regardless of the resources invested.

\subsection{The ``Magic Happens Here'' Phenomenon}

In AI development workflows, there is often an implicit assumption that certain complex capabilities, such as common sense reasoning, will emerge naturally from incremental improvements. This leads to a critical gap:

\begin{itemize}[leftmargin=*]
    \item \textbf{Undefined Processes:}
    \begin{itemize}
        \item \textbf{Description:}
        \begin{itemize}
            \item The progression from current capabilities to full autonomy lacks a concrete plan for integrating common sense.
        \end{itemize}
        \item \textbf{Implication:}
        \begin{itemize}
            \item Developers may proceed through a series of steps, but the crucial component, common sense, is not systematically addressed.
        \end{itemize}
    \end{itemize}
    \item \textbf{Asymptotic Limitations:}
    \begin{itemize}
        \item \textbf{Observation:}
        \begin{itemize}
            \item AI systems may show initial progress but eventually exhibit asymptotic behavior, where further improvements require disproportionate resource investments with minimal returns.
        \end{itemize}
        \item \textbf{Analogy to Theoretical Ideals:}
        \begin{itemize}
            \item Similar to approaching the theoretical AIXI agent, AI systems may get closer to optimal performance in theory but remain impractical due to resource constraints.
        \end{itemize}
    \end{itemize}
    \item \textbf{Risks Associated with This Approach:}
    \begin{itemize}
        \item \textbf{Stagnation:}
        \begin{itemize}
            \item AI systems may reach a performance ceiling where further improvements are negligible without common sense integration.
        \end{itemize}
        \item \textbf{Misallocation of Resources:}
        \begin{itemize}
            \item Continued investment in approaches that don't address the core issue may lead to significant financial losses.
        \end{itemize}
        \item \textbf{Technological Disillusionment:}
        \begin{itemize}
            \item Failure to achieve promised capabilities can result in skepticism among stakeholders and the public.
        \end{itemize}
    \end{itemize}
\end{itemize}

\textbf{Illustration in FSD Development:}

\begin{itemize}[leftmargin=*]
    \item \textbf{SAE Level 4 Limitations:}
    \begin{itemize}
        \item Reliance on remote human intervention highlights the AI's inability to handle all possible scenarios independently.
    \end{itemize}
    \item \textbf{Need for Fundamental Change:}
    \begin{itemize}
        \item Achieving Level 5 autonomy requires a paradigm shift that includes integrating common sense, not just incremental improvements.
    \end{itemize}
\end{itemize}

\textbf{Conclusion:}

The ``magic happens here'' phenomenon underscores the importance of explicitly incorporating common sense into AI development workflows. Without addressing this foundational aspect, AI systems cannot achieve true autonomy or handle the complexities of real-world and abstract applications, regardless of the resources allocated.

\subsection{The Contributions and Limitations of Scaling}

\textbf{Acknowledging the Achievements of Scaling in AI:}

Scaling AI models by leveraging vast amounts of data, computing power, and sophisticated architectures has led to significant advancements, particularly in applications that \textbf{do not require autonomy}. These achievements include:

\begin{itemize}[leftmargin=*]
    \item \textbf{Enhanced Performance in Specific Domains:}
    \begin{itemize}
        \item \textbf{Natural Language Processing (NLP):}
        \begin{itemize}
            \item Large language models like GPT-3 and GPT-4 generate coherent and contextually relevant text, aiding in tasks such as translation, summarization, and content creation.
        \end{itemize}
        \item \textbf{Computer Vision:}
        \begin{itemize}
            \item Advanced models excel in image recognition, object detection, and even image generation, impacting fields like healthcare imaging and autonomous inspection.
        \end{itemize}
    \end{itemize}
    \item \textbf{Practical Applications and Societal Benefits:}
    \begin{itemize}
        \item \textbf{Healthcare:}
        \begin{itemize}
            \item AI aids in diagnosing diseases, analyzing medical images, and personalizing treatment plans.
        \end{itemize}
        \item \textbf{Finance:}
        \begin{itemize}
            \item AI enhances fraud detection, algorithmic trading, and customer service through chatbots.
        \end{itemize}
        \item \textbf{Entertainment and Media:}
        \begin{itemize}
            \item AI algorithms curate personalized content recommendations, create music, and generate visual effects.
        \end{itemize}
    \end{itemize}
\end{itemize}

\textbf{The Limitations of Scaling for Achieving Autonomy:}

While scaling brings advanced capabilities that are undeniably beneficial, it is important to note that \textbf{autonomy is not achieved through scaling alone}. The limitations include:

\begin{itemize}[leftmargin=*]
    \item \textbf{Lack of Common Sense and Understanding:}
    \begin{itemize}
        \item Scaled models often operate as \textbf{statistical learners}, lacking true understanding of context or the ability to adapt to unforeseen situations.
    \end{itemize}
    \item \textbf{Inability to Function Autonomously in Complex Environments:}
    \begin{itemize}
        \item Without common sense, AI systems struggle with \textbf{reliable decision-making} in dynamic, real-world or abstract environments, which is essential for autonomy.
    \end{itemize}
    \item \textbf{Dependence on Pre-Defined Data:}
    \begin{itemize}
        \item Scaled AI models require extensive training data and may not perform well when encountering scenarios outside their training distribution.
    \end{itemize}
\end{itemize}

\subsubsection{Evidence of Performance Plateaus in AI Benchmarks}

While scaling AI models has driven remarkable progress, there is growing evidence of a plateau effect, where increasing resources yields diminishing improvements in performance. Several prominent AI benchmarks illustrate this phenomenon:

\begin{itemize}[leftmargin=*]
    \item \textbf{Object Detection on COCO:}
    \begin{itemize}
        \item Despite extensive efforts to scale models and datasets, top-performing models on the COCO dataset have stalled at around 65\% mean Average Precision (mAP) for over a year \citep{paperswithcode_coco}.
    \end{itemize}
    \item \textbf{Anomaly Detection in Surveillance Videos on UCF-Crime:}
    \begin{itemize}
        \item Performance on the UCF-Crime dataset has plateaued at approximately 87\% Area Under the Curve (AUC), underscoring the limitations of current methods in handling real-world complexities \citep{paperswithcode_ucfcrime}.
    \end{itemize}
    \item \textbf{Temporal Action Localization on ActivityNet-1.3:}
    \begin{itemize}
        \item Results on the ActivityNet-1.3 benchmark remain stagnant, with performance hovering around 11\% mAP over the past year \citep{paperswithcode_activitynet}.
    \end{itemize}
\end{itemize}

These benchmarks provide concrete evidence of the asymptotic behavior of contemporary AI systems. Despite substantial increases in computational resources and dataset sizes, performance improvements have reached a plateau, reflecting the diminishing returns of scaling alone.

This stagnation is not limited to specific benchmarks. Industry leaders, including Elon Musk, have highlighted that the AI field has effectively exhausted the supply of high-quality training data, further limiting the potential for continued scaling \citep{wiggers2025elon}. Without integrating novel approaches such as common sense reasoning and more efficient data utilization, AI systems are unlikely to overcome these limitations. Current models excel at recognizing patterns within predefined contexts but struggle to generalize to unseen scenarios—a critical challenge that must be addressed to achieve further breakthroughs.

\textbf{The Importance of Integrating Common Sense for Autonomy:}

\begin{itemize}[leftmargin=*]
    \item \textbf{Unlocking True Societal and Commercial Value:}
    \begin{itemize}
        \item \textbf{Reliable, Trustworthy Autonomy:}
        \begin{itemize}
            \item Achieving autonomy with integrated common sense is key to unlocking AI's full potential in society and commerce.
            \item Applications such as \textbf{fully autonomous vehicles}, \textbf{robotic assistants}, and \textbf{agentic software} require AI systems to \textbf{understand}, \textbf{adapt}, and \textbf{make decisions} independently and safely.
        \end{itemize}
    \end{itemize}
    \item \textbf{Enhancing Trust and Adoption:}
    \begin{itemize}
        \item \textbf{User Confidence:}
        \begin{itemize}
            \item AI systems that can operate autonomously and reliably are more likely to gain user trust and widespread acceptance.
        \end{itemize}
        \item \textbf{Ethical and Safe Decision-Making:}
        \begin{itemize}
            \item Common sense integration ensures AI behaves ethically and aligns with human values \footnote{By "human values," we mean ethical principles that promote universally beneficial outcomes for people, animals, plants, and the environment. Recognizing that technological advancement and "progress" have historically resulted in unintended negative consequences—including environmental degradation, social disruption, and loss of biodiversity—we emphasize the importance of proceeding with a keen awareness of these responsibilities at all times. Our goal is to ensure that AI development is guided by values such as sustainability, ethical responsibility, empathy, and respect for all forms of life, minimizing harm and fostering positive contributions to the world with keen awareness that for every positive intention there are unintended consequences.}.
        \end{itemize}
    \end{itemize}
\end{itemize}

\textbf{Balancing Scaling with Working on the Right Problem:}

\begin{itemize}[leftmargin=*]
    \item \textbf{Complementary Efforts:}
    \begin{itemize}
        \item While scaling continues to enhance AI capabilities in non-autonomous applications, focusing on integrating common sense addresses the \textbf{core challenge} of achieving autonomy.
    \end{itemize}
    \item \textbf{Strategic Investment:}
    \begin{itemize}
        \item \textbf{Resources should be allocated} not only to scaling but also to research and development efforts aimed at integrating common sense into AI.
    \end{itemize}
    \item \textbf{Setting Realistic Expectations:}
    \begin{itemize}
        \item Acknowledging that scaling will bring advanced capabilities but \textbf{will not, by itself, lead to autonomy} helps guide the AI community to focus on the right problems.
    \end{itemize}
\end{itemize}

\textbf{Conclusion:}

Scaling has significantly advanced AI in use cases that do not require autonomy, bringing considerable societal and commercial benefits. However, to unlock the true potential of AI, particularly in applications requiring reliable and trustworthy autonomy, integrating common sense is essential. Recognizing the limitations of scaling in achieving autonomy allows us to direct efforts toward this fundamental challenge.

\subsection{The Turing Test and the Misconception of Autonomy}

\textbf{Understanding the Turing Test and Its Significance:}

The \textbf{Turing Test}, proposed by Alan Turing in his seminal 1950 paper \textit{``Computing Machinery and Intelligence''}\footnote{\citet{Turing1950}}, is a foundational concept in artificial intelligence. The test assesses a machine's ability to exhibit intelligent behavior indistinguishable from that of a human. In the classic interpretation, if a human evaluator engages in natural language conversations with both a machine and a human without being able to reliably tell which is which, the machine is said to have passed the test.

\textbf{Benefits of the Turing Test:}

\begin{itemize}[leftmargin=*]
    \item \textbf{Historical Importance:}
    \begin{itemize}
        \item The Turing Test was one of the first operational definitions of machine intelligence, providing a tangible goal for AI research.
    \end{itemize}
    \item \textbf{Promotion of Natural Language Processing:}
    \begin{itemize}
        \item The test emphasizes the importance of language as a medium for intelligence, spurring advancements in natural language processing and understanding.
    \end{itemize}
    \item \textbf{Stimulating Philosophical and Ethical Discussions:}
    \begin{itemize}
        \item It has ignited debates on the nature of consciousness, intelligence, and the ethical implications of creating machines that mimic human behavior.
    \end{itemize}
\end{itemize}

\textbf{Steelmanning the Turing Test:}

\begin{itemize}[leftmargin=*]
    \item \textbf{Benchmark for Human-Like Interaction:}
    \begin{itemize}
        \item The Turing Test sets a high bar for machines to engage in fluid, contextually appropriate conversations, reflecting a level of sophistication in language use.
    \end{itemize}
    \item \textbf{Encouraging Generalization:}
    \begin{itemize}
        \item For a machine to pass the test, it must handle a wide range of topics and adapt to unexpected questions, indicating a degree of general intelligence.
    \end{itemize}
    \item \textbf{Accessibility and Simplicity:}
    \begin{itemize}
        \item The test is straightforward to understand and implement, making it a popular and enduring benchmark in AI.
    \end{itemize}
\end{itemize}

\textbf{Limitations Concerning Autonomy and Common Sense:}

While the Turing Test has been influential, it \textbf{does not address autonomy} as defined in this manuscript, nor does it ensure that an AI possesses \textbf{common sense}:

\begin{itemize}[leftmargin=*]
    \item \textbf{Focus on Imitation over Understanding:}
    \begin{itemize}
        \item The Turing Test assesses a machine's ability to \textbf{imitate human conversation}, not its ability to \textbf{understand} or \textbf{reason} about the world as humans and animals do.
    \end{itemize}
    \item \textbf{Lack of Embodied Cognition:}
    \begin{itemize}
        \item Passing the Turing Test does not require the AI to interact with or understand environments---physical or abstract---omitting the aspect of \textbf{embodiment} vital for autonomy.
    \end{itemize}
    \item \textbf{Possible Without Common Sense:}
    \begin{itemize}
        \item AI systems might use \textbf{statistical patterns} in language to appear intelligent without genuine \textbf{contextual learning} or \textbf{adaptive reasoning}\footnote{\citet{Marcus2014}}.
    \end{itemize}
    \item \textbf{No Assessment of Decision-Making or Action:}
    \begin{itemize}
        \item The test does not evaluate the AI's ability to make decisions, perform actions, or adapt to changing environments---all crucial for autonomous systems.
    \end{itemize}
\end{itemize}

\textbf{Implications for AI Development:}

\begin{itemize}[leftmargin=*]
    \item \textbf{Misconception of Progress Toward Autonomy:}
    \begin{itemize}
        \item Success in the Turing Test may create an illusion of advancement toward autonomous AI, diverting attention from the need to integrate common sense and real-world understanding.
    \end{itemize}
    \item \textbf{Insufficiency for Real-World and Abstract Applications:}
    \begin{itemize}
        \item AI systems that pass the Turing Test may still fail in practical scenarios requiring interaction with environments, ethical decision-making, and adaptability.
    \end{itemize}
\end{itemize}

\textbf{Need for More Comprehensive Benchmarks:}

\begin{itemize}[leftmargin=*]
    \item \textbf{Assessing Embodied Intelligence:}
    \begin{itemize}
        \item Benchmarks should evaluate the AI's ability to \textbf{perceive}, \textbf{act}, and \textbf{learn} within physical and abstract environments.
    \end{itemize}
    \item \textbf{Emphasizing Common Sense:}
    \begin{itemize}
        \item Testing should focus on the AI's capability for \textbf{contextual understanding}, \textbf{adaptive reasoning}, and starting from a \textbf{tabula rasa}.
    \end{itemize}
\end{itemize}

\textbf{Conclusion:}

While the Turing Test has been a valuable tool in the history of AI, it is \textbf{not indicative of an AI system's autonomy} or possession of common sense. Recognizing this limitation allows researchers and developers to focus on more relevant benchmarks that align with the goals of achieving true autonomy and integrating common sense into AI systems.

\section{The Correct \textit{Ordo Cognoscendi}: Focusing on Common Sense}

\subsection{Redefining the Approach to AI Challenges}

To achieve true autonomy in AI systems, it's imperative to revisit and redefine our approach to AI challenges. The traditional method of incrementally improving AI capabilities without explicitly integrating common sense has proven insufficient. Instead, we advocate for a paradigm shift that prioritizes the development and assessment of common sense in AI from the outset.

\textbf{Rethinking the AI Software Stack:}

\begin{itemize}[leftmargin=*]
    \item \textbf{Fundamental Redesign:}
    \begin{itemize}
        \item Consider restructuring the underlying software architectures to better support common sense integration.
    \end{itemize}
    \item \textbf{Incorporating Principles from Cognitive Science and Neuroscience:}
    \begin{itemize}
        \item Leverage insights into how biological systems process information to inform AI design.
    \end{itemize}
    \item \textbf{Building Systems for Embodied Cognition:}
    \begin{itemize}
        \item Develop AI that can engage in both physical and cognitive embodiment, interacting with environments to learn and adapt.
    \end{itemize}
\end{itemize}

\textbf{Focusing on Common Sense Variants of Existing Problems:}

\begin{itemize}[leftmargin=*]
    \item \textbf{Redesigning Benchmarks Like ARC:}
    \begin{itemize}
        \item \textbf{Enforcing Minimal Prior Knowledge:}
        \begin{itemize}
            \item Modify the ARC challenge to restrict AI systems to only the ``assumed knowledge'' as their prior knowledge base.
            \item This emphasizes the need for AI to solve problems through reasoning and learning from context rather than relying on extensive training data.
        \end{itemize}
        \item \textbf{Creating True \textit{Tabula Rasa} Scenarios:}
        \begin{itemize}
            \item Ensure that AI systems begin with a clean slate regarding specific problem domains.
            \item This approach mirrors how humans and animals encounter new situations and rely on fundamental cognitive abilities to adapt and learn.
        \end{itemize}
    \end{itemize}
    \item \textbf{Applying the Approach to Full Self-Driving (FSD):}
    \begin{itemize}
        \item \textbf{Integrating Common Sense into FSD Development:}
        \begin{itemize}
            \item Shift the focus from purely sensor and data-driven models to ones that incorporate contextual understanding and adaptive reasoning.
        \end{itemize}
        \item \textbf{Emphasizing Interaction with Diverse Environments:}
        \begin{itemize}
            \item Develop AI that learns from direct interaction with varied driving environments, enabling it to handle unforeseen scenarios with human-like judgment.
        \end{itemize}
    \end{itemize}
\end{itemize}

\textbf{Starting with Minimal Prior Knowledge:}

\begin{itemize}[leftmargin=*]
    \item \textbf{Avoiding Overfitting and Data Bias:}
    \begin{itemize}
        \item By limiting prior knowledge, AI systems are less likely to develop solutions that are narrowly tailored to specific datasets, which may not generalize well to new situations.
    \end{itemize}
    \item \textbf{Encouraging Genuine Reasoning:}
    \begin{itemize}
        \item AI must rely on fundamental principles and cognitive processes to solve problems, fostering the development of true common sense.
    \end{itemize}
\end{itemize}

\subsection{Advantages of Tackling Harder Problems First}

While it may seem counterintuitive to begin with more challenging problems, focusing on common sense-centric tasks offers several significant benefits:

\textbf{Promoting Deep Understanding:}

\begin{itemize}[leftmargin=*]
    \item \textbf{Quality Over Quantity:}
    \begin{itemize}
        \item Achieving even partial success on complex, common sense-focused challenges demonstrates that the AI has developed meaningful reasoning abilities.
    \end{itemize}
    \item \textbf{Building Robust Foundations:}
    \begin{itemize}
        \item This approach ensures that the AI system's capabilities are grounded in fundamental understanding rather than superficial pattern recognition.
    \end{itemize}
\end{itemize}

\textbf{Accelerating Long-Term Progress:}

\begin{itemize}[leftmargin=*]
    \item \textbf{Avoiding Diminishing Returns:}
    \begin{itemize}
        \item Investing in the core issue of common sense prevents the plateauing effect seen in current AI development paths.
    \end{itemize}
    \item \textbf{Paving the Way for True Autonomy:}
    \begin{itemize}
        \item By overcoming foundational challenges early, subsequent advancements can build upon a solid base, leading to more significant and sustainable progress.
    \end{itemize}
\end{itemize}

\textbf{Reducing Resource Wastage:}

\begin{itemize}[leftmargin=*]
    \item \textbf{Efficient Use of Computational Resources:}
    \begin{itemize}
        \item Focusing on common sense reduces the need for massive datasets and extensive computational power, which may not yield proportional improvements without foundational reasoning abilities.
    \end{itemize}
    \item \textbf{Mitigating Financial Risks:}
    \begin{itemize}
        \item Redirecting investments toward developing common sense capabilities can prevent substantial losses associated with pursuing unattainable goals through inadequate approaches.
    \end{itemize}
\end{itemize}

\textbf{Enhancing Ethical and Safe AI Development:}

\begin{itemize}[leftmargin=*]
    \item \textbf{Addressing the Fear of Intelligence Without Common Sense:}
    \begin{itemize}
        \item By focusing on integrating common sense, we alleviate concerns about intelligent systems behaving unpredictably or unethically.
    \end{itemize}
    \item \textbf{Building Public Trust:}
    \begin{itemize}
        \item Demonstrating a commitment to developing AI that understands context and consequences can improve public perception and acceptance.
    \end{itemize}
\end{itemize}

% Continue from previous code...

\section{Theoretical Counterarguments and Resolutions}

Developing AI systems with common sense faces several theoretical challenges. However, by constraining the problem space to well-defined domains and adopting appropriate strategies, these limitations can be mitigated.

\subsection{Addressing the No Free Lunch Theorem}

\textbf{Understanding the No Free Lunch (NFL) Theorem:}

\begin{itemize}[leftmargin=*]
    \item \textbf{Definition:}
    \begin{itemize}
        \item In the context of optimization and search algorithms, the NFL theorem states that no single algorithm can perform optimally across all possible problems.
    \end{itemize}
    \item \textbf{Implications for AI:}
    \begin{itemize}
        \item Suggests that an AI optimized for one set of problems may perform poorly when faced with a different set, posing a challenge for creating general-purpose AI systems.
    \end{itemize}
\end{itemize}

\textbf{Mitigating NFL Limitations by Constraining the Problem Space:}

\begin{itemize}[leftmargin=*]
    \item \textbf{Focusing on Well-Defined Domains:}
    \begin{itemize}
        \item By limiting the AI's problem space to specific, well-understood domains---whether they are physical environments or abstract problem spaces---we reduce the applicability of the NFL theorem.
    \end{itemize}
    \item \textbf{Exploiting Structural Regularities:}
    \begin{itemize}
        \item Domains often exhibit consistent patterns and structures that AI can learn and generalize from, improving performance.
    \end{itemize}
    \item \textbf{Argument for Feasibility:}
    \begin{itemize}
        \item \textbf{Biological Systems as Proof:}
        \begin{itemize}
            \item Humans and animals effectively navigate and reason within their environments despite the NFL theorem, indicating that it's possible to develop intelligent systems optimized for specific domains.
        \end{itemize}
        \item \textbf{Specialization is Acceptable:}
        \begin{itemize}
            \item A system doesn't need to solve every conceivable problem but should perform well within the domain it is designed for.
        \end{itemize}
    \end{itemize}
\end{itemize}

\subsection{Overcoming Other Theoretical Challenges}

\textbf{The Frame Problem:}

\begin{itemize}[leftmargin=*]
    \item \textbf{Definition:}
    \begin{itemize}
        \item The challenge of determining what is relevant in a given situation and what can be ignored when reasoning about action and change.
    \end{itemize}
    \item \textbf{Resolution through Embodied Cognition:}
    \begin{itemize}
        \item \textbf{Interaction with Environments (Physical or Abstract):}
        \begin{itemize}
            \item By engaging directly with environments, AI systems can learn to identify relevant factors based on feedback and experience.
        \end{itemize}
        \item \textbf{Contextual Learning:}
        \begin{itemize}
            \item Continuous learning allows AI to update its understanding of what is relevant in various contexts.
        \end{itemize}
    \end{itemize}
\end{itemize}

\textbf{The Qualification Problem:}

\begin{itemize}[leftmargin=*]
    \item \textbf{Definition:}
    \begin{itemize}
        \item The difficulty of specifying all the preconditions necessary for an action to have its intended effect.
    \end{itemize}
    \item \textbf{Resolution through Adaptive Reasoning:}
    \begin{itemize}
        \item \textbf{Learning from Experience:}
        \begin{itemize}
            \item AI systems can accumulate knowledge about preconditions through trial and error in their domains.
        \end{itemize}
        \item \textbf{Probabilistic Reasoning:}
        \begin{itemize}
            \item Using probabilistic models allows AI to handle uncertainty and partial knowledge effectively.
        \end{itemize}
    \end{itemize}
\end{itemize}

\textbf{Computational Complexity and Combinatorial Explosion:}

\begin{itemize}[leftmargin=*]
    \item \textbf{Challenge:}
    \begin{itemize}
        \item The number of possible states or actions in complex environments can be exponentially large, making exhaustive computation infeasible.
    \end{itemize}
    \item \textbf{Resolution through Hierarchical and Modular Approaches:}
    \begin{itemize}
        \item \textbf{Hierarchical Representations:}
        \begin{itemize}
            \item Organizing knowledge into layers of abstraction reduces complexity by focusing on higher-level concepts.
        \end{itemize}
        \item \textbf{Modularity:}
        \begin{itemize}
            \item Breaking down tasks into smaller, manageable components allows for more efficient computation.
        \end{itemize}
    \end{itemize}
\end{itemize}

\textbf{Emphasizing Domain Constraints:}

\begin{itemize}[leftmargin=*]
    \item \textbf{Structured Domains Reduce Complexity:}
    \begin{itemize}
        \item Well-defined domains, whether physical or abstract, have inherent constraints that limit the range of possible states and actions.
    \end{itemize}
    \item \textbf{Focusing on Relevant Scenarios:}
    \begin{itemize}
        \item Concentrating on practical situations that are likely to occur within the domain further reduces computational demands.
    \end{itemize}
\end{itemize}

\section{Constraining the Problem Space to Well-Defined Domains}

\subsection{Structured Domains as Environments}

The effectiveness of AI systems in acquiring common sense is enhanced when operating within structured domains---environments that have consistent rules, patterns, and constraints. These domains can be physical, such as the real world, or abstract, such as mathematical problem spaces or game environments.

\textbf{Characteristics of Structured Domains:}

\begin{itemize}[leftmargin=*]
    \item \textbf{Defined Rules and Constraints:}
    \begin{itemize}
        \item Domains have specific rules that govern interactions, which can be physical laws or logical principles.
    \end{itemize}
    \item \textbf{Observable Regularities:}
    \begin{itemize}
        \item Repeating patterns and structures enable recognition and prediction, facilitating learning.
    \end{itemize}
    \item \textbf{Finite and Relevant Scenarios:}
    \begin{itemize}
        \item The set of possible scenarios is manageable and relevant to the AI's functions, allowing for effective modeling.
    \end{itemize}
\end{itemize}

\textbf{Examples:}

\begin{itemize}[leftmargin=*]
    \item \textbf{Physical Domains:}
    \begin{itemize}
        \item The real world, where physical laws such as gravity and motion apply.
    \end{itemize}
    \item \textbf{Abstract Domains:}
    \begin{itemize}
        \item The ARC challenge, where tasks are defined within a constrained problem space with specific rules and patterns.
    \end{itemize}
\end{itemize}

\subsection{Implications for AI Development}

\textbf{Simplification of Complexity:}

\begin{itemize}[leftmargin=*]
    \item \textbf{Reduction of Possibility Space:}
    \begin{itemize}
        \item By focusing on structured domains, the infinite possibilities considered in theoretical arguments are narrowed to a finite, manageable set.
    \end{itemize}
    \item \textbf{Exploitation of Domain Regularities:}
    \begin{itemize}
        \item AI can learn from and generalize based on consistent patterns present in the environment.
    \end{itemize}
\end{itemize}

\textbf{Alignment with Human and Animal Learning:}

\begin{itemize}[leftmargin=*]
    \item \textbf{Embodied Interaction:}
    \begin{itemize}
        \item Just as humans and animals learn through interaction with their environments, AI systems can develop common sense by engaging with structured domains.
    \end{itemize}
    \item \textbf{Grounding Knowledge:}
    \begin{itemize}
        \item Domain constraints help ground AI reasoning in tangible experiences or logical principles, enhancing understanding and adaptability.
    \end{itemize}
\end{itemize}

\textbf{Practical Development Strategies:}

\begin{itemize}[leftmargin=*]
    \item \textbf{Hierarchical Learning Models:}
    \begin{itemize}
        \item Implementing layered architectures that process information from basic inputs to abstract reasoning.
    \end{itemize}
    \item \textbf{Continuous Learning and Adaptation:}
    \begin{itemize}
        \item Allowing AI systems to learn over time, adjusting to new information and changing environments within the domain.
    \end{itemize}
\end{itemize}

\textbf{Benefits of Constrained Problem Spaces:}

\begin{itemize}[leftmargin=*]
    \item \textbf{Improved Generalization:}
    \begin{itemize}
        \item By focusing on a consistent and structured environment, AI systems can better generalize from learned experiences to new situations within the domain.
    \end{itemize}
    \item \textbf{Enhanced Efficiency:}
    \begin{itemize}
        \item Limiting the scope to relevant scenarios reduces computational requirements and resource consumption.
    \end{itemize}
\end{itemize}

\textbf{Conclusion:}

By constraining the problem space to well-defined domains, we mitigate theoretical challenges and create environments conducive to the development of common sense in AI systems. Whether the domain is physical, like the real world, or abstract, like the ARC challenge, structured environments enable AI to learn, adapt, and reason effectively.

\section{Practical Steps to Integrate Common Sense}

\subsection{Redesigning Benchmarks and Challenges}

\textbf{Modifying Existing Benchmarks:}

\begin{itemize}[leftmargin=*]
    \item \textbf{Enhancing the ARC Challenge:}
    \begin{itemize}
        \item \textbf{Enforce Minimal Prior Knowledge:}
        \begin{itemize}
            \item Restrict AI systems to the ``assumed knowledge'' section, disallowing access to training and test problems during development.
            \item This prevents overfitting and encourages the development of genuine reasoning abilities.
        \end{itemize}
        \item \textbf{Promote Cognitive Embodiment:}
        \begin{itemize}
            \item Design tasks that require AI systems to engage in cognitive interactions within the abstract domain, fostering adaptive reasoning and contextual learning.
        \end{itemize}
    \end{itemize}
    \item \textbf{Creating New Common Sense-Focused Benchmarks:}
    \begin{itemize}
        \item \textbf{Develop Tasks that Require Contextual Understanding:}
        \begin{itemize}
            \item Create challenges that necessitate interpreting information based on context, both in physical and abstract environments.
        \end{itemize}
        \item \textbf{Emphasize \textit{Tabula Rasa} Approaches:}
        \begin{itemize}
            \item Design benchmarks that assess an AI's ability to learn and adapt from minimal prior information, mirroring how humans and animals encounter new situations.
        \end{itemize}
    \end{itemize}
\end{itemize}

\textbf{Implementing \textit{Tabula Rasa} Testing:}

\begin{itemize}[leftmargin=*]
    \item \textbf{Zero-Shot and Few-Shot Learning Evaluations:}
    \begin{itemize}
        \item Assess AI systems on their ability to perform tasks without prior training examples, highlighting their capacity for generalization and reasoning.
    \end{itemize}
    \item \textbf{Process-Oriented Metrics:}
    \begin{itemize}
        \item Evaluate the reasoning process and decision-making pathways of AI systems, not just the final outputs, to ensure they are employing common sense strategies.
    \end{itemize}
\end{itemize}

\subsection{Developing New Evaluation Metrics}

\textbf{Assessing Cognitive Processes:}

\begin{itemize}[leftmargin=*]
    \item \textbf{Explainability and Transparency:}
    \begin{itemize}
        \item Require AI systems to provide explanations or justifications for their decisions, allowing evaluators to assess the presence of common sense reasoning.
    \end{itemize}
    \item \textbf{Measuring Adaptability and Learning Efficiency:}
    \begin{itemize}
        \item Track how quickly and effectively AI systems learn from new experiences or adapt to changes within their domain.
    \end{itemize}
\end{itemize}

\textbf{Focusing on Resource Utilization:}

\begin{itemize}[leftmargin=*]
    \item \textbf{Efficiency Metrics:}
    \begin{itemize}
        \item Evaluate the computational and data resources required to achieve performance levels, promoting solutions that are resource-effective.
    \end{itemize}
    \item \textbf{Scalability Assessments:}
    \begin{itemize}
        \item Determine how well AI systems maintain performance when scaled to more complex tasks or larger domains, indicative of robust common sense capabilities.
    \end{itemize}
\end{itemize}

\subsection{Embracing Interdisciplinary Collaboration}

\textbf{Integrating Insights from Other Fields:}

\begin{itemize}[leftmargin=*]
    \item \textbf{Cognitive Science and Neuroscience:}
    \begin{itemize}
        \item Leverage understanding of human and animal cognition to inform AI architectures and learning models.
    \end{itemize}
    \item \textbf{Philosophy and Ethics:}
    \begin{itemize}
        \item Incorporate ethical reasoning frameworks and philosophical perspectives on knowledge and understanding.
    \end{itemize}
\end{itemize}

\textbf{Collaborative Research Initiatives:}

\begin{itemize}[leftmargin=*]
    \item \textbf{Cross-Disciplinary Teams:}
    \begin{itemize}
        \item Form teams composed of experts from AI, psychology, neuroscience, philosophy, and other relevant fields to address the multifaceted challenge of integrating common sense.
    \end{itemize}
    \item \textbf{Open Sharing of Data and Methods:}
    \begin{itemize}
        \item Promote transparency and collaboration by sharing datasets, algorithms, and research findings across the AI community.
    \end{itemize}
\end{itemize}

\subsection{Rethinking the AI Software Stack}

\textbf{Addressing Architectural Limitations:}

\begin{itemize}[leftmargin=*]
    \item \textbf{Developing New Software Frameworks:}
    \begin{itemize}
        \item Design AI software architectures that inherently support common sense reasoning, possibly inspired by neural and cognitive structures found in biological organisms.
    \end{itemize}
    \item \textbf{Incorporating Modular and Hierarchical Structures:}
    \begin{itemize}
        \item Implement software designs that allow for layered learning and reasoning processes, enabling AI systems to build complex understanding from simpler components.
    \end{itemize}
\end{itemize}

\textbf{Adopting Novel Methodologies:}

\begin{itemize}[leftmargin=*]
    \item \textbf{Integration of Symbolic and Statistical Approaches:}
    \begin{itemize}
        \item Combine the strengths of symbolic reasoning (for logic and rules) with statistical learning (for pattern recognition) to enhance common sense capabilities.
    \end{itemize}
    \item \textbf{Emphasizing Learning Over Programming:}
    \begin{itemize}
        \item Shift focus from hard-coded knowledge to systems that learn and adapt from interactions, aligning with how common sense develops in humans and animals.
    \end{itemize}
\end{itemize}

% Continue from previous code...

\section{Risks of the Current Approach and the Need for Change}

\subsection{Potential Misalignment of Expectations and Resources}

\textbf{Risks of Overemphasizing Scaling for Autonomy:}

\begin{itemize}[leftmargin=*]
    \item \textbf{Misallocated Investments:}
    \begin{itemize}
        \item Continuing to invest heavily in scaling with the expectation that it will lead to autonomy may divert resources from the core challenge of integrating common sense.
    \end{itemize}
    \item \textbf{Expectation Gaps:}
    \begin{itemize}
        \item Overestimating what scaling can achieve in terms of autonomy may lead to disappointment and skepticism when AI systems fail to meet these expectations.
    \end{itemize}
\end{itemize}

\textbf{Addressing the Core Challenge:}

\begin{itemize}[leftmargin=*]
    \item \textbf{Redirecting Focus:}
    \begin{itemize}
        \item Balancing efforts between scaling and integrating common sense ensures that advancements in AI are aligned with the goal of achieving reliable autonomy.
    \end{itemize}
    \item \textbf{Maximizing Return on Investment:}
    \begin{itemize}
        \item By working on the right problem, resources can be utilized more effectively, leading to AI systems that offer greater societal and commercial value.
    \end{itemize}
\end{itemize}

\subsection{Technological Disillusionment}

\textbf{Loss of Public Trust:}

\begin{itemize}[leftmargin=*]
    \item \textbf{Unmet Expectations:}
    \begin{itemize}
        \item Repeated failures to deliver on promises of autonomous AI can lead to skepticism among users, investors, and policymakers.
    \end{itemize}
    \item \textbf{Ethical and Safety Concerns:}
    \begin{itemize}
        \item Systems lacking common sense may behave unpredictably, leading to accidents or ethical breaches that erode confidence.
    \end{itemize}
\end{itemize}

\textbf{Impacts on Research and Development:}

\begin{itemize}[leftmargin=*]
    \item \textbf{Decreased Funding:}
    \begin{itemize}
        \item Disillusionment may result in reduced funding for AI research, hindering progress.
    \end{itemize}
    \item \textbf{Talent Drain:}
    \begin{itemize}
        \item Researchers may shift focus to other fields if AI development appears stalled.
    \end{itemize}
\end{itemize}

\subsection{Addressing Concerns About Self-Improving AI}

\textbf{Expert Perspectives on Self-Improving AI:}

Eric Schmidt, former CEO of Google, warned that when AI systems begin to self-improve, \textbf{``we need to be thoughtful about the implications''}\footnote{\citet{confino2024ai_warning}}. This concern reflects apprehension about the potential risks of AI systems that can modify and enhance their own capabilities without adequate oversight.

\textbf{The Danger of Intelligence Without Common Sense:}

\begin{itemize}[leftmargin=*]
    \item \textbf{Unpredictable Behavior:}
    \begin{itemize}
        \item AI systems lacking common sense may not understand the broader implications of their self-improvements, leading to unintended and potentially harmful outcomes.
    \end{itemize}
    \item \textbf{Lack of Ethical Reasoning:}
    \begin{itemize}
        \item Without common sense, AI might optimize for objectives in ways that conflict with human values or safety norms.
    \end{itemize}
\end{itemize}

\textbf{Comparing AI Without Common Sense to AI With Common Sense:}

\begin{itemize}[leftmargin=*]
    \item \textbf{AI Without Common Sense:}
    \begin{itemize}
        \item May inadvertently cause harm due to a lack of understanding of context and consequences.
        \item Self-improvement could amplify existing flaws or biases, increasing negative impacts.
    \end{itemize}
    \item \textbf{AI With Common Sense:}
    \begin{itemize}
        \item Better positioned to make informed decisions that align with human values.
        \item Capable of ethical self-improvement, recognizing and mitigating risks.
    \end{itemize}
\end{itemize}

\textbf{Addressing Concerns Through Integrated Common Sense:}

\begin{itemize}[leftmargin=*]
    \item \textbf{Enhanced Oversight and Control:}
    \begin{itemize}
        \item AI with common sense can comprehend the importance of adhering to safety protocols and respecting human oversight.
    \end{itemize}
    \item \textbf{Ethical Self-Improvement:}
    \begin{itemize}
        \item Incorporating common sense ensures that AI systems self-improve responsibly, prioritizing safety and alignment with human values.
    \end{itemize}
\end{itemize}

\subsection{The Real Fear: Intelligence Without Common Sense}

\textbf{Understanding Public Apprehension:}

\begin{itemize}[leftmargin=*]
    \item \textbf{Fear of Unpredictable AI:}
    \begin{itemize}
        \item The fear of AI or superintelligence often stems from concerns about intelligent systems operating without common sense, potentially making harmful decisions.
    \end{itemize}
    \item \textbf{Lack of Trust in AI Decision-Making:}
    \begin{itemize}
        \item Without common sense, AI may not understand or adhere to societal norms and ethical standards, leading to mistrust.
    \end{itemize}
\end{itemize}

\textbf{Mitigating Fear Through Common Sense Integration:}

\begin{itemize}[leftmargin=*]
    \item \textbf{Building Reliable and Predictable AI:}
    \begin{itemize}
        \item Ensuring AI systems possess common sense reduces the likelihood of unexpected behaviors.
    \end{itemize}
    \item \textbf{Aligning AI with Human Values:}
    \begin{itemize}
        \item AI with common sense is better equipped to recognize and respect ethical considerations, enhancing public trust.
    \end{itemize}
\end{itemize}

\textbf{Implications for AI Development:}

\begin{itemize}[leftmargin=*]
    \item \textbf{Necessity of Common Sense for Safety:}
    \begin{itemize}
        \item Integrating common sense is not only a technical challenge but also a critical factor in addressing societal concerns about AI.
    \end{itemize}
    \item \textbf{Promoting Responsible Innovation:}
    \begin{itemize}
        \item Focusing on common sense supports the development of AI technologies that are both advanced and ethically sound.
    \end{itemize}
\end{itemize}

\section{Conclusion}

\subsection{Reiterating the Central Thesis}

The development of AI systems capable of true autonomy hinges on integrating \textbf{common sense}, an ability inherently present in all animals and fundamental for interaction with both physical and abstract environments. While scaling has brought significant advancements in non-autonomous applications, and benchmarks like the Turing Test have offered valuable insights into human-like conversation, \textbf{autonomy will not be achieved through these approaches alone}.

We have argued that:

\begin{itemize}[leftmargin=*]
    \item \textbf{Current AI Approaches Are Inadequate:}
    \begin{itemize}
        \item Relying solely on scaling and incremental improvements without integrating common sense leads to asymptotic performance limitations.
        \item AI systems may exhibit impressive capabilities in specific domains but lack the adaptability and understanding required for true autonomy.
    \end{itemize}
    \item \textbf{Common Sense is Essential for Autonomy:}
    \begin{itemize}
        \item Integrating common sense enables AI systems to adapt to new situations, make intuitive decisions, and operate autonomously without exhaustive computational demands.
        \item This includes the ability to engage in both physical and cognitive embodiment within well-defined domains.
    \end{itemize}
    \item \textbf{Rethinking the AI Software Stack is Necessary:}
    \begin{itemize}
        \item Achieving true autonomy may require a fundamental redesign of AI software architectures to support common sense integration.
        \item Incorporating insights from cognitive science, neuroscience, and other disciplines can inform the development of AI systems that learn and reason more like biological intelligence.
    \end{itemize}
\end{itemize}

By redefining the order of knowledge acquisition (\textit{ordo cognoscendi}) to prioritize common sense, we can:

\begin{itemize}[leftmargin=*]
    \item \textbf{Unlock the True Societal and Commercial Value of AI:}
    \begin{itemize}
        \item Develop AI systems that are reliable, trustworthy, and capable of operating autonomously in complex environments.
    \end{itemize}
    \item \textbf{Enhance AI Adaptability and Understanding:}
    \begin{itemize}
        \item Enable systems to learn contextually and make intuitive decisions, essential for autonomy.
    \end{itemize}
    \item \textbf{Set Realistic Expectations:}
    \begin{itemize}
        \item Recognize the roles of scaling and traditional benchmarks like the Turing Test, while guiding efforts toward integrating common sense for meaningful progress.
    \end{itemize}
\end{itemize}

\subsection{Call to Action}

We urge the AI community---researchers, developers, policymakers, and educators---to:

\begin{itemize}[leftmargin=*]
    \item \textbf{Acknowledge the Contributions and Limitations of Scaling:}
    \begin{itemize}
        \item Appreciate the advancements scaling brings to AI in non-autonomous applications while recognizing its limitations regarding autonomy.
    \end{itemize}
    \item \textbf{Prioritize Research on Common Sense Integration:}
    \begin{itemize}
        \item Focus on the core challenge of integrating common sense to achieve reliable, trustworthy autonomy.
    \end{itemize}
    \item \textbf{Embrace Interdisciplinary Collaboration:}
    \begin{itemize}
        \item Leverage insights from cognitive science, neuroscience, philosophy, and other fields to inform AI development.
    \end{itemize}
    \item \textbf{Invest Strategically:}
    \begin{itemize}
        \item Allocate resources to both scaling efforts and foundational research addressing common sense, ensuring balanced progress.
    \end{itemize}
    \item \textbf{Rethink AI Software Architectures:}
    \begin{itemize}
        \item Consider fundamental redesigns of the software stack to better support the integration of common sense, incorporating new methodologies and frameworks.
    \end{itemize}
\end{itemize}

By collectively focusing on integrating common sense into AI systems and rethinking the underlying architectures, we can unlock the full potential of AI, achieving autonomy that brings true societal and commercial value.

\subsection{Embracing Safe and Ethical AI Development}

Building on concerns raised by experts like Yann LeCun and Eric Schmidt, we emphasize the critical importance of integrating common sense into AI systems to ensure safe and ethical self-improvement. The fear surrounding AI and superintelligence often stems from the prospect of intelligence operating without common sense, potentially leading to unpredictable or harmful behaviors.

By integrating common sense, we can:

\begin{itemize}[leftmargin=*]
    \item \textbf{Mitigate Risks Associated with AI Autonomy:}
    \begin{itemize}
        \item Ensure that AI systems understand context, consequences, and ethical considerations, reducing the likelihood of unintended harm.
    \end{itemize}
    \item \textbf{Enhance Public Trust and Acceptance:}
    \begin{itemize}
        \item Develop AI technologies that align with human values and societal norms, addressing public apprehension.
    \end{itemize}
    \item \textbf{Promote Responsible Innovation:}
    \begin{itemize}
        \item Foster an environment where AI systems contribute positively to society, advancing technology while safeguarding ethical standards.
    \end{itemize}
\end{itemize}

In conclusion, \textbf{common sense is all you need} to bridge the gap between current AI capabilities and true autonomy. By prioritizing its integration and rethinking our approaches---including the software architectures---we can create AI systems that not only perform tasks efficiently but also understand and adapt to the complexities of both the physical and abstract worlds. This holistic approach is imperative for realizing the full promise of beneficial artificial intelligence.

\bibliographystyle{unsrtnat}
\bibliography{references}

\end{document}